\pdfoutput=1

\documentclass[11pt]{article}

\usepackage[]{EACL2023}

\usepackage{times}
\usepackage{latexsym}

\usepackage[T1]{fontenc}

\usepackage[utf8]{inputenc}

\usepackage{microtype}

\usepackage{inconsolata}

\usepackage{times}

\usepackage{soul}
\usepackage{url}
\usepackage{graphicx}
\usepackage{amsmath}
\usepackage{stackrel,amssymb}
\usepackage{booktabs}
\usepackage{balance}
\usepackage{graphics}
\usepackage{graphicx}
\usepackage{multirow}
\usepackage{caption}

\usepackage{makecell}
\newcommand*{\scale}[2][4]{\scalebox{#1}{$#2$}}

\title{A Survey of Multi-task Learning in Natural Language Processing: \\ Regarding Task Relatedness and Training Methods}

\author{{\bf Zhihan Zhang, Wenhao Yu, Mengxia Yu, Zhichun Guo, Meng Jiang} \\
University of Notre Dame, Notre Dame, IN, USA \\
\normalsize{ {\tt \{zzhang23, wyu1, myu2, zguo5, mjiang2\}@nd.edu}}
}

\begin{document}
\maketitle


\begin{abstract}
Multi-task learning (MTL) has become increasingly popular in natural language processing (NLP) because it improves the performance of related tasks by exploiting their commonalities and differences. Nevertheless, it is still not understood very well how multi-task learning can be implemented based on the relatedness of training tasks. In this survey, we review recent advances of multi-task learning methods in NLP, with the aim of summarizing them into two general multi-task training methods based on their task relatedness: (i) joint training and (ii) multi-step training. We present examples in various NLP downstream applications, summarize the task relationships and discuss future directions of this promising topic.

\end{abstract}

\vspace{0.05in}
\section{Introduction}
\label{sec:intro}
Machine learning generally involves training a model to perform a single task.
By focusing on one task, the model ignores knowledge from the training signals of \emph{related tasks} \cite{ruder2017overview}.
There are a great number of tasks in NLP, from syntax parsing to information extraction, from machine translation to question answering: each requires a model dedicated to learning from data. Biologically, humans learn natural languages, from basic grammar to complex semantics in a single brain \cite{hashimoto2017joint}. In the field of machine learning, multi-task learning (MTL) aims to leverage useful information shared across multiple related tasks to improve the generalization performance on all tasks~\cite{caruana1997multitask}. 
In deep neural networks, it is generally achieved by sharing part of hidden layers between different tasks, while keeping several task-specific output layers.
MTL offers advantages like improved data efficiency, reduced overfitting, and fast learning by leveraging auxiliary information \cite{crawshaw2020multi}.

There have been relevant surveys that looked into architecture designs and optimization algorithms in MTL. \citet{ruder2017overview} classified different MTL frameworks into two categories: hard parameter sharing and soft parameter sharing, as well as some earlier MTL examples in both non-neural and neural models;
\citet{zhang2018overview} expanded such two ``how to share'' categories into five categories, including feature learning approach, low-rank approach, task clustering approach, task relation learning approach, and decomposition approach;
\citet{crawshaw2020multi} presented more recent models in both single-domain and multi-modal architectures, as well as an overview of optimization methods in MTL.
Nevertheless, it is still not clearly understood \emph{how} to design and train a single model to handle a variety of NLP tasks according to \textbf{task relatedness}.
Especially when faced with a set of tasks that are seldom simultaneously trained previously, it is of crucial importance that researchers find proper auxiliary tasks and assess the feasibility of such multi-task learning attempt.

\begin{figure}[t]
    \centering
    \includegraphics[width=0.5\textwidth]{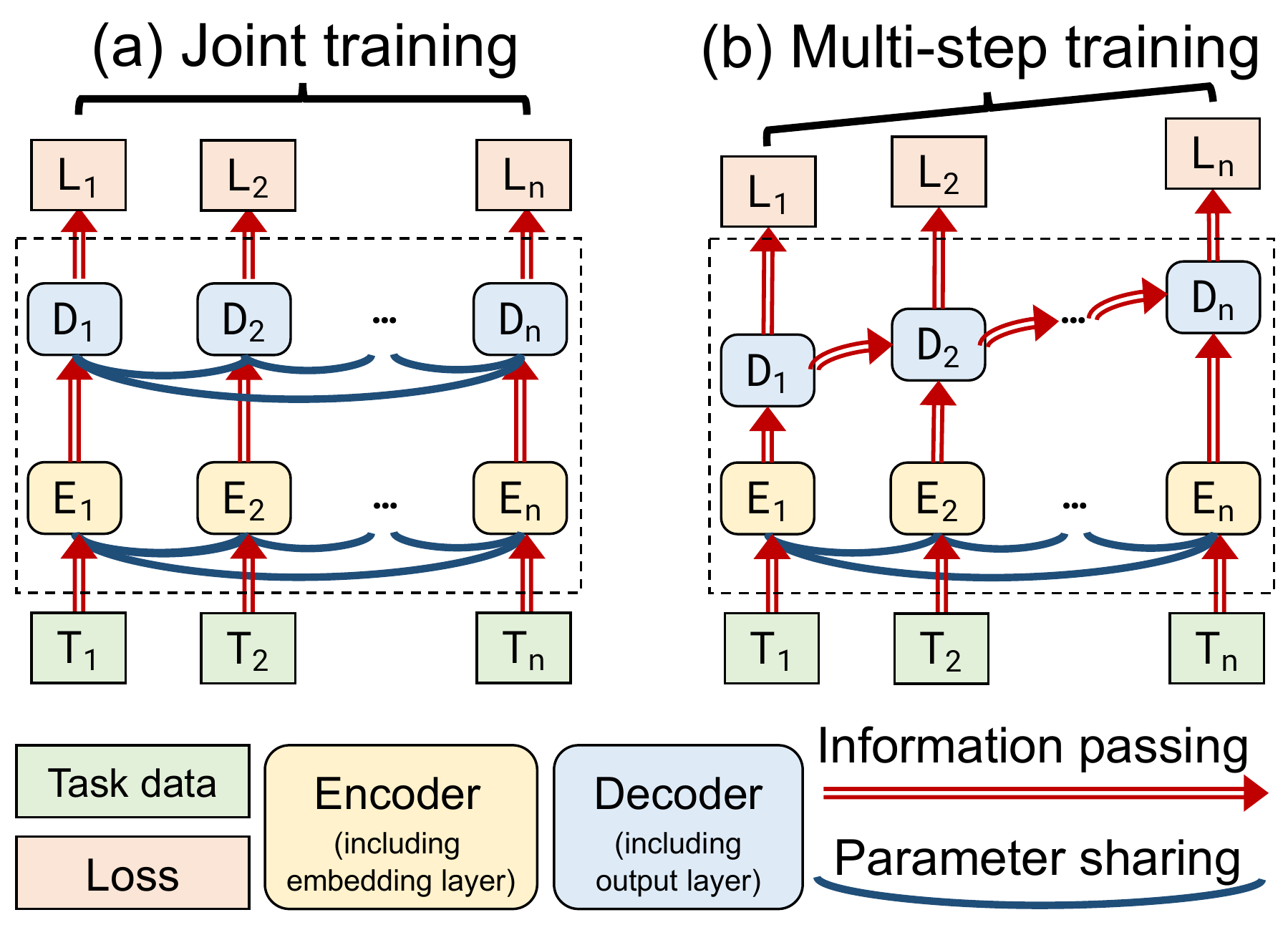}
    \label{fig:framework-joint}
    \vspace{-0.1in}
    \caption{Two multi-task learning frameworks.}
    \label{fig:frameworks-basic}
    \vspace{-0.2in}
\end{figure}

\begin{table*}[t]
\begin{center}
\scale[0.71]{\begin{tabular}{l|l|l|l}
\toprule
\textbf{Multi-task training} & \textbf{Multi-task frameworks} & \textbf{Multi-task frameworks} & \multirow{2}*{\textbf{Related papers}} \\ 
\textbf{methods in our survey} & \textbf{defined in \citet{ruder2017overview}} & \textbf{defined in \citet{crawshaw2020multi}} & \\
\midrule
{\multirow{5}*{\makecell[c]{Joint Training }}} & Deep relationship network & Shared trunk & \cite{liu2015representation,dong2015multi} \\
&  Cross-stitch network & Cross-talk & \cite{liu2016recurrent,xiao2018gated} \\
&   Weighting losses with uncertainty &  & \cite{xiong2018dcn+,xu2019multi}\\
&   Sluice network & & \cite{ruder2017sluice} \\
&   & Adversarial feature separation & \cite{liu2017adversarial,mao2020adaptive} \\
\midrule
{\multirow{2}*{\makecell[c]{Multi-step Training}}} & & Prediction distillation & \cite{dinan2019wizard,lewis2020retrieval} \\
&  Low supervision & Cascaded information & \cite{sogaard2016deep,hashimoto2017joint} \\
\bottomrule
\end{tabular}}
\end{center}
\vspace{-0.15in}
\caption{Categories of multi-task learning frameworks in two related surveys can be merged into our proposed joint training and multi-step training frameworks.}
\vspace{-0.15in}
\label{tab:frameworks}
\end{table*}

In this paper, we first review recent approaches on multi-task training methods in popular NLP applications.
We find that these approaches can be categorized into \textit{two multi-task training methods} according to the types of task relatedness: (i) joint training methods and (ii) multi-step training methods.
\textbf{Joint training} is commonly used when all given tasks can be performed simultaneously and all the task-specific data can be learned simultaneously.
In joint training, model parameters are shared (either via soft or hard parameter sharing\footnote{We do not specifically distinguish different parameter sharing designs, since this topic is not the focus of our survey. We refer readers to learn details in \citet{ruder2017overview}.}) among encoders and decoders so that the tasks can be jointly trained to benefit from shared representations, as shown in Figure \ref{fig:frameworks-basic}(a).
In contrast, \textbf{multi-step training} is used when some task's input needs to be determined by the outputs or hidden representations of previous task(s).
Due to such task dependencies, the task-specific decoders are connected as a multi-step path starting from the encoder ``node'', as shown in Figure \ref{fig:frameworks-basic}(b).

{Therefore, different from previous surveys which focus on architecture designs (\textit{e.g.}, how to share parameters in \cite{ruder2017overview} and \cite{zhang2018overview}) and optimization methods (\textit{e.g.}, loss weighting and regularization in \cite{crawshaw2020multi}), our motivation lies in categorizing two major multi-task training methods in NLP, according to \textbf{task relatedness}. In fact, task relatedness is the key to determine what training method to use, then the training method decides the general scope of available architecture designs. With specific application tasks, readers are able to identify the ideal training method from our review before looking for detailed module design or loss optimization in previous surveys. We also show that how the MTL techniques covered in previous surveys can be matched with the two training methods in Table \ref{tab:frameworks}.}

The remainder of this survey is organized as follows.
Section \ref{sec:model} includes an overview of MTL models in NLP and the rationales of using MTL. 
Section \ref{sec:application} presents a number of joint and multi-step training applications in different fields of NLP.
Section \ref{sec:task} analyzes the task relatedness involved in these MTL approaches.
Section \ref{sec:future} discusses future directions. Section \ref{sec:conclusion} concludes the paper.

\section{Multi-task Training Methods}
\label{sec:model}
\subsection{Encoder-Decoder Architecture and Two Multi-Task Training Frameworks}
\label{sec:framework}
Suppose we train a model on $n$ NLP tasks $T_1$, $\cdots$, $T_n$ on a dataset $\mathcal{D} = \{(X^{(i)}, Y^{(i)})\}{|}^{N}_{i=1}$ with $N$ data points.
For the $j$-th NLP task, the model is trained with $\{(X^{(i)}_j, Y^{(i)}_j)\}{|}^{N}_{i=1}$, where $X^{(i)}_j$ is a component of the input $X^{(i)}$, and $Y^{(i)}_j$ is the desired output.
The input components of different tasks can be the same, but the desired outputs are usually different.
We formulate the multi-task frameworks discussed in this paper under the popular encoder-decoder architecture which are mainly composed of three components: (a) the encoder layer (including the embedding layer), (b) the decoder layer (including the output layer for classification or generation), and (c) loss and optimization.


\vspace{0.05in}
\noindent\textbf{Encoder layer.}
In NLP networks, an embedding layer is usually applied to generate the embedding vectors of the basic elements of the input $X^{(i)}$ 
For the $j$-th task, the encoder layer learns the hidden state of $X^{(i)}_j$ as a vector $\mathbf{h}^{(i)}_j$:
\vspace{-0.25cm}
\begin{equation}
    \mathbf{h}^{(i)}_j = \mathrm{Encoder}(X^{(i)}_j, \Theta_{E_j}),
    \vspace{-0.25cm}
\end{equation}
where $\Theta_{E_j}$ denotes the parameters of $j$-th task's encoder. 
Parameters of different encoders can be shared.
Popular encoder modules include BiLSTM and BERT~\cite{devlin2019bert}. 

\vspace{0.05in}
\noindent\textbf{Decoder layer.}
When the tasks are \textit{independent} with each other at decoding, the decoder of the $j$-th task transforms the hidden state into an output: 
\vspace{-0.25cm}
\begin{equation}
    \hat{Y}^{(i)}_j = \mathrm{Decoder}_j(\mathbf{h}^{(i)}_j, \Theta_{D_j}).
    \label{eq:decoder_joint}
    \vspace{-0.25cm}
\end{equation}
When the tasks are \textit{sequentially dependent}, the decoder of the $j$-th task needs the output of the $(j-1)$-th task, then we have
\vspace{-0.25cm}
\begin{align}
    \hat{Y}^{(i)}_j = \mathrm{Decoder}_j(\hat{Y}^{(i)}_{j-1}, \mathbf{h}^{(i)}_j, \Theta_{D_j}).
    \label{eq:decoder_multihop}
    \vspace{-0.25cm}
\end{align}
where $\Theta_{D_j}$ denotes the parameters of $j$-th task's decoder. Practically, $\hat{Y}^{(i)}_{j-1}$ are often presented as hidden representations of the decoder prediction to enable end-to-end training.
Parameters of different decoders can be shared.
Popular decoder choices include MLP, LSTM and the Transformer decoder.

According to the two types of task dependencies, the multi-task learning frameworks define and organize the decoders in two different ways. 
As shown in Figure~\ref{fig:frameworks-basic}, (i) the \textbf{joint training} framework is for the tasks that are independent at decoding; and (ii) the \textbf{multi-step training} framework is for tasks that are sequentially dependent. It can be easily generalized when the task dependencies form a directed acyclic graph, in which sequential dependence is a special and common case.

\vspace{0.05in}
\noindent\textbf{Optimization.} 
A common optimization approach of MTL is to optimize the weighted sum of loss functions from different tasks, (i.e., $\text{Loss} = \lambda_j \sum_{j=1}^n \text{Loss}_j$)
then compute the gradient descent to update all trainable parameters ($\{\Theta_{E_j}\}_{j=1}^n$, $\{\Theta_{D_j}\}_{j=1}^n$). 
The weights of $\{\lambda_j\}_{j=1}^n$ can be either pre-defined or dynamically adjusted~\cite{kendall2018uncertainty, xiong2018dcn+}. It is worth mentioning that optimization in MTL has many alternative ways. For example, \citet{sogaard2016deep} choose a random task $t$ from a pre-defined task sets to optimize its loss at each iteration. {Readers can find a more detailed review of MTL optimization methods in \citet{crawshaw2020multi}, which is not the main focus of this paper.}




\subsection{How does MTL Work}
One of the prerequisites of multi-task learning is the relatedness among different tasks and their data. 
Most work prefers to train positively correlated tasks in a multi-task setting.
Such tasks have similar objectives or relevant data, and can boost each other to form consistent predictions through shared lower-level representations. 
According to \citet{caruana1997multitask}, in MTL, tasks prefer hidden representations that other tasks prefer. MTL enables shared representations to include features from all tasks, thus improving the consistency of task-specific decoding in each sub-task. 
Furthermore, the co-existence of features from different objectives naturally performs feature crosses, which enables the model to learn more complex features. 

According to the experiments by \citet{standley2020tasks}, tasks are more likely to benefit from MTL when using a larger network. This can be achieved as the emergence of deep neural frameworks in recent years. Many deep models, like BERT \cite{devlin2019bert} and T5 \cite{raffel2020t5}, have strong generalization ability to fit a variety of tasks with minute changes. Therefore, different tasks can be learned through similar models, especially in the field of NLP where the encoder-decoder architecture has become a norm.

With the above premises, deep models are able to benefit from MTL in multiple perspectives. First, MTL improves data efficiency for each sub-task. Different tasks provide different aspects of information, enriching the expression ability of the hidden representation to the input text. Besides, different tasks have different noise patterns, which acts as an implicit data augmentation method. This encourages the multi-task model to produce more generalizable representations in shared layers. Thus, the model is prevented from overfitting to a single task and gains stronger generalization ability, which helps itself perform well when faced with new tasks from a similar environment \cite{baxter2000model}.
Multi-task learning are also effective for low-resource tasks \cite{lin2018multi, johnson2017google}. Co-training with a high-resource task in a similar domain, low-resource tasks receive ampler training signals which prevents the model from overfitting on the limited data.


Auxiliary tasks in MTL can serve as conditions or hints for the main task. Such setting usually falls into the category of \textit{multi-step training}. Providing additional conditions reduces the distribution space of possible outputs, thus lower the prediction difficulty of the main task. Such conditions can serve as additional features during decoding, including external knowledge pieces, low-level NLP tasks (\textit{e.g.}, part-of-speech tagging or syntactic parsing) or relevant snippets extracted from long documents. When some features are difficult for the main task to directly learn, explicit supervision signals of such features, if available, enables the model to ``eavesdrop", \textit{i.e.}, obtaining these features through the learning of auxiliary task \cite{ruder2017overview}. 

\label{sec:need}

\section{Training Methods: Applications}
\label{sec:application}
\subsection{Joint Training Applications}
\label{sec:joint}

\begin{figure*}[th]
    \centering
    \includegraphics[width=\textwidth]{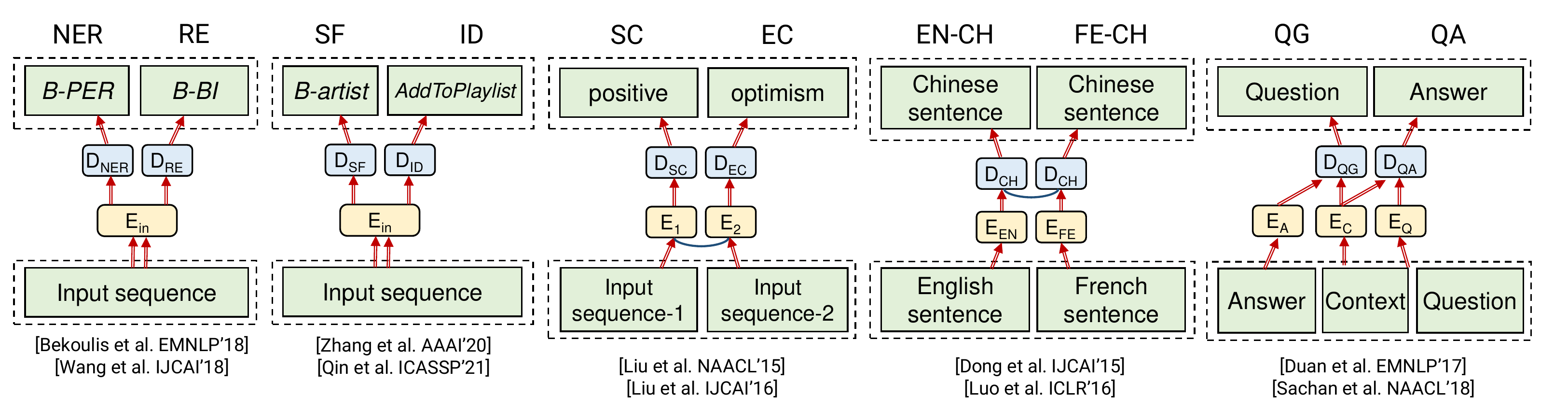}
    \vspace{-0.3in}
    \caption{Five joint training NLP applications that have been discussed from $\S$\ref{sec:IE} to $\S$\ref{sec:NLG}.}
    \vspace{-0.11in}
    \label{fig:joint-application}
\end{figure*}



In this section, we list a series of recent approaches of joint training in different fields of NLP (shown in Figure \ref{fig:joint-application}), including information extraction, spoken language understanding, text classification, machine translation and language generation.

\vspace{-0.05in}
\subsubsection{Information Extraction (IE)}
\label{sec:IE}

Two popular tasks that are usually jointly performed in IE are named entity recognition (NER) and relation extraction (RE).
NER seeks to locate and classify named entities in text into pre-defined categories, such as names and locations.
NER is often tackled by sequence labeling methods, or say, token-wise classifiers.
RE aims to extract semantic relationships between two or more entities, and there are multiple ways to define the RE task in the multi-task training approach.

First, \citet{zhou2019joint} predicted the type of relation mentioned in a sentence by the RE decoder. It works for simple sentences such as those that have a pair of entities and one type of relation, e.g., ``[President Obama] was \emph{born in} [Honolulu].'' However, one sentence may have multiple types of relations. 
Second, \citet{zheng2017joint} predicted a relation tag for every pair of tokens. If the decoder performs perfectly, it can identify any number and any types of relations in a sentence. However, the complexity is too high to be effectively trained with annotated data.
Third, \citet{bekoulis2018adversarial,wang2018joint-2} treated RE as a sequence labeling problem. So both NER and RE decoders are token-wise classifiers.
As shown in Figure \ref{fig:joint-application}, for example, $B$-$BI$ tag represents the beginning word of subject entity (person) or object entity (location) in the ``born\_in'' ($BI$) relation. Therefore, if multiple tag sequences can be generated, they can identify any number, and any type of relations in the input sentence.

\vspace{-0.05in}
\subsubsection{Spoken Language Understanding}
\label{sec:SLU}

Spoken Language Understanding (SLU) plays an important role in spoken dialogue system~\cite{SLU_survey}.
SLU aims at extracting the semantics from user utterances, which is a critical component of task-oriented dialogue.
Concretely, it captures semantic constituents of the utterance and identifies the user's intent.
These two tasks are typically known as slot filling (SF) and intent detection (ID), respectively. Each word in the utterance corresponds to one slot label, and a specific intent is assigned to the whole utterance.
An example of these two sub-tasks is given below:
\vspace{-0.2cm}
\begin{table}[h]
\begin{center}
\scale[0.82]{\begin{tabular}{|ccccccc|}
\hline
\textbf{Word} & Put & Kanye & into & my & rap & playlist \\
& $\downarrow$ & $\downarrow$ & $\downarrow$ & $\downarrow$ & $\downarrow$ & $\downarrow$ \\
\textbf{Slot} & O & B-artist & O & O & B-playlist & O \\
\textbf{Intent} & \multicolumn{6}{l|}{AddToPlaylist} \\
\hline
\end{tabular}}
\end{center}
\end{table}
\vspace{-0.3cm}

\noindent  Since two sub-tasks share the same input utterance, they usually share a single utterance encoder and are jointly trained~\cite{liu2016attention, castellucci2019multilingual}. Recent state-of-the-art SLU models build bi-directional interactions during encoding ~\cite{liu2019cmnet, zhang2020graph, qin2021cointeractive}. Therefore, two tasks mutually impact each other before making respective predictions. It is worth noting that there is also a line of work that uses the hidden states of intent detection to assist slot filling~\cite{goo2018slot, qin2019stack, qin2021glgin}. This can be considered as a combination of joint training and multi-step training: intent detection helps the prediction of slot filling, but finally their predictions are integrated to perform the parent (larger) SLU task.

\vspace{-0.05in}
\subsubsection{Sentence/Document Classification}
\label{sec:SC}

Sentence/document classification is one of the fundamental tasks in NLP with broad applications such as sentiment classification (SC), emotion classification (EC), and stance detection.
However, the construction of large-scale high-quality datasets is extremely labor-intensive. Therefore, multi-task learning plays an important role in leveraging potential correlations among related classification tasks 
to extract common features, increase corpus size implicitly and yield classification improvements. 
Popular multi-task learning setting in text classification has two categories. 
First, one dataset is annotated with multiple labels and one input is associated with multiple outputs~\cite{liu2015representation,yu2018improving, gui2020multi}.
Second, multiple datasets have their respective labels, i.e., multiple inputs with multiple outputs, where samples from different tasks are jointly learned in parallel~\cite{liu2016recurrent,liu2017adversarial}. 
Most existing work leverages joint training for different sentence/document classification tasks. Specifically, \citet{liu2016recurrent} proposed three different parameter sharing designs under the joint training framework, and further compared their performances.

\vspace{-0.05in}
\subsubsection{Multilinguality}
\label{sec:MT}

Languages differ lexically but are closely related on the semantic and/or the syntactic levels. Such correlation
across different languages motivates the multi-task learning on multilingual data. Neural machine translation (NMT) is the most important application. \citet{dong2015multi} first proposed a multi-task learning framework based on Seq2Seq to conduct NMT from one source language to multiple target languages. \citet{luong2015multi} extended it with many-to-one and many-to-many approaches. Many-to-one is useful for translating multi-source languages to the target language, in which only the decoder is shared. Many-to-many studies the effect of unsupervised translation between multiple languages. 
\citet{zhu2019ncls} proposed to improve cross-lingual summarization by jointly training with monolingual summarization and machine translation. \citet{arivazhagan2019massively} built a massive multi-lingual translation model handling 103 languages, and conducted experiments on multiple sampling schema for building joint training dataset. 

{Besides, unlabelled data from the target language is also a common source of multi-task cross-lingual training. \citet{ahmad2019cross} collected unannotated sentences from auxiliary languages to assist learning language-agnostic representations. \citet{van2021masked} incorporated a masked language modeling objective using unlabeled data from target language to perform zero-shot transfer.}

\begin{figure*}[th]
    \centering
    \includegraphics[width=0.9\textwidth]{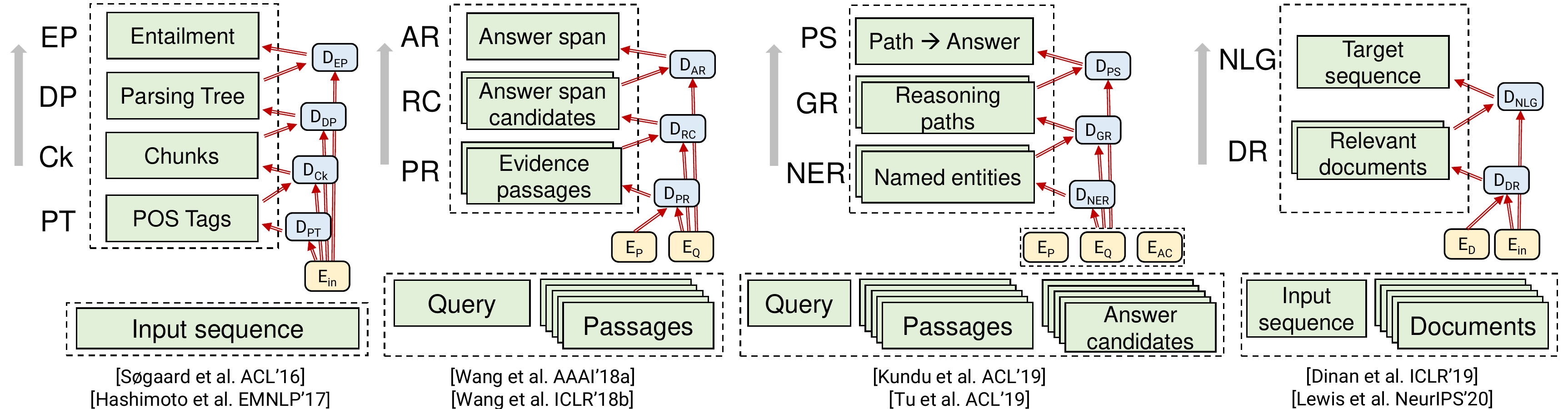}
    \caption{Four multi-step training NLP applications discussed at $\S$\ref{sec:MLU}, $\S$\ref{sec:MPQA} (2nd and 3rd subfigures) and $\S$\ref{sec:RAG}.}
    \vspace{-0.15in}
    \label{fig:multistep-app}
\end{figure*}

\vspace{-0.05in}
\subsubsection{Natural Language Generation (NLG)}
\label{sec:NLG}

Recent success in deep generative modeling have led to significant advances in NLG, motivated by an increasing need to understand and derive meaning from language~\cite{yu2020survey}.
The relatedness between different generation tasks promotes the application of multi-task learning in NLG.

For example, \citet{guo2018soft} proposed to jointly learn abstractive summarization (AS) and question generation (QG).
An accurate summary of a document is supposed to contain all its salient information. This goal is consistent with that of question generation (QG), which looks for salient questioning-worthy details. 
Besides, QG and question answering (QA) are often trained as dual tasks.
\citet{tang2017question} proposed a joint learning framework that connects QG and QA. QA improves QG through measuring the relevance between the generated question and the answer.
QG improves QA by providing additional signal which stands for the probability of generating a question given the answer.
A similar framework was also employed in \citet{duan2017question,sachan2018self}.

In other applications, semantic parsing is gaining attention for knowledge-based question answering since it does not rely on hand-crafted features.  \cite{shen2019multi} developed a joint learning approach where a pointer-equipped semantic parsing model is designed to resolve coreference in conversations, and naturally empower joint learning with a novel type-aware entity detection model. Researchers also found NLU tasks, \textit{e.g.}, input meaning representation learning~\cite{qader2019semi} or entity mention prediction~\cite{dong2020injecting}, can improve the performance of generating sentences

{Multi-view learning is also applied in NLG approaches for auxiliary learning objectives. Input data are erased partially to create distinct views, and divergence metrics are usually learned along with the main loss to force the model generate consistent predictions across different views of the same input. Typical approaches include \citet{clark2018semi} which built up the multi-view learning paradigm in IE and NLG tasks. In addition, \citet{shen2020simple} upgraded the network by combining multiple cutoff methods to create augmented data, and achieved success in translation tasks.}

\subsection{Multi-step Training: Applications}
\label{sec:multihop}
We list recent approaches of multi-step training in different field of NLP (as shown in Figure \ref{fig:multistep-app}), such as language understanding, multi-passage question answering and natural language 
generation. 

\vspace{-0.05in}
\subsubsection{Multi-level Language Understanding}
\label{sec:MLU}

The potential for leveraging multiple levels of representations has been demonstrated in various ways in the field of NLP. For example, Part-Of-Speech (POS) tags are used for syntactic parsers. The parsers are used
to improve higher-level tasks, such as natural language inference.
\citet{sogaard2016deep} showed when learning POS tagging and chunking, it is consistently better to have POS supervision at the innermost rather than the outermost layer.
\citet{hashimoto2017joint} predicted increasingly complex NLP tasks at successively deeper layers for POS tagging, chunking, dependency parsing, semantic relatedness, and textual entailment, by considering linguistic hierarchies. Lower level predictions may influence predictions in higher levels, \textit{e.g.}, if the semantic relatedness between two sentences is very low, they are unlikely to entail each other. 
Similar architecture can be found in \citet{sanh2019hierarchical}, where low-level tasks are name entity recognition and entity mention detection, with coreference resolution and relation extraction supervising at higher levels.

\vspace{-0.05in}
\subsubsection{Multi-Passage Question Answering}
\label{sec:MPQA}

Question answering (QA) models may need to construct answers by querying multiple passages (\textit{e.g.}, paragraphs, documents).
Given a question, multi-passage QA (MPQA) requires AI models identify an answer span from multiple evidence passages. Due to the complexity of MPQA, it is usually achieved by multiple sub-tasks. Thus, multi-step training is utilized by many approaches in MPQA.



Typically, MPQA can be split into a 3-phase task.
\emph{Passage retrieval} (PR) is to select relevant evidence passages according to the question.
\emph{Reading comprehension} (RC) is to extract multiple answer span candidates from the retrieved set of relevant passages.
\emph{Answer reranking} (AR) is to re-score multiple answer candidates based on the question and evidence passages.
There exist dependencies between these tasks: evidence passages are generated by PR and fed into RC as input; the answer span candidates are generated by RC and given into AR as input.
So, as shown in Figure~\ref{fig:multistep-app}, the decoders form a multi-hop path starting from the shared encoder.
\citet{hu2019retrieve} proposed a typical approach called RE${}^{3}$ (for REtriever, REader, and REranker).
The retriever used TF-IDF cosine similarities to prune irrelevant passages.
The reader is a token classifier that predicts the start and end indices of answer candidates per segment.
The reranker prunes redundant span candidates and then predict the reranking scores.
Other works in MPQA also considered 2-phase approaches, such as PR+RC \cite{wang2018r3} or RC+AR \cite{wang2018joint}, which are simplified versions of the above framework.
Similar approaches have been developed for many domains such as news \cite{nishida2018retrieve}, and web questions \cite{lin2018denoising}.



Another branch of this task is \textit{multi-choice} MPQA, where we have a set of answer candidates for the given question.
\cite{kundu2019exploiting} proposed to exploit explicit paths for multi-hop reasoning over structured knowledge graphs. The model attempted to extract implicit relations from text through entity pair representations, and compose them to encode each path. It composed the passage representations along each path to compute a passage-based representation. Then it can explain the reasoning via these explicit paths through the passages. The sub-tasks are \emph{named entity recognition} (NER), \emph{graph-based reasoning} (GR) to extract and encode paths, and passage-based \emph{path scoring} (PS).
So, the multi-task QA systems perform interpretable and accurate reasoning~\cite{welbl2018constructing,tu2019multi}.

\vspace{-0.04in}
\subsubsection{Retrieval-augmented Text Generation}
\vspace{-0.02in}
\label{sec:RAG}

In NLG, the input sequence alone often contains limited knowledge to support neural generation models to produce the desired output, so the performance of generation is still far from satisfactory in many real-world scenarios~\cite{yu2020survey}. 
Retrieval-augmented generation models use the input sequence to retrieve relevant information (e.g., a background document) and use it as additional contexts when generating the target sequence. For example, \citet{dinan2019wizard} proposed to tackle the knowledge-aware dialogue by first selecting knowledge from a large pool of document candidates and generate a response based on the selected knowledge and context. 
To enhance the aforementioned idea, \citet{kim2020sequential} presented a sequential latent variable model to keep track of the prior and posterior distribution over knowledge. It not only reduced the ambiguity caused from the diversity in knowledge selection of conversation but also better leveraged the response information for proper choice of knowledge. 
Similar retrieval-augmented generation approaches have been applied in question generation~\cite{lewis2020retrieval}, comment generation~\cite{lin2019learning}, image captioning~\cite{xu2019unified}, summarization~\cite{cao2018retrieve}, and long form QA~\cite{krishna2021hurdles}.

\section{NLP Task Relatedness}
\vspace{-0.02in}
\label{sec:task}

In this section, we summarize the characteristics of the aforementioned MTL approaches, and look into the task relatedness between the sub-tasks.


\vspace{-0.05in}
\subsection{Joint Training }

\paragraph{Joint training with similar tasks.}
Joint training with a similar task is the classical choice for multi-task learning. According to \citet{caruana1997multitask}, more similar tasks share more hidden units. Hence, similar tasks are more likely to benefit from shared generic representations.
However, what kind of tasks can be considered as ``similar" are not always evident in the deep learning era. 
Empirically selecting similar tasks is still the most mainstream method~\cite{ruder2017overview,worsham2020multi}.
To get some intuitions what a similar task can be, here we introduce some prominent examples.
\cite{dong2015multi} proposed training neural machine translation from one language into multiple languages simultaneously;
\cite{yu2018improving} proposed a joint training framework for sentiment classification and emotion classification;
\cite{guo2018soft} proposed abstractive summarization can be jointly learned with question generation.
\cite{yang2019knowledge} jointly trained question categorization and answer retrieval.

Recently, \citet{aribandi2021ext5} attempted to empirically select
a set of tasks (from 107 NLP tasks) to transfer from, using a multi-task objective of mixing supervised tasks with self-supervised objectives for language model pre-training. 
Some recent work also tried to select appropriate sub-tasks based on manually defined features~\cite{lin2019choosing,sun2021ranking}.
In addition, \citet{guo2019autosem} used multi-armed bandits to select tasks and a Gaussian Process to control the mixing rates. \citet{gradts} further utilized the attention-heads importance distribution of the Transformer as a criterion to select auxiliary tasks. Aside from NLP, \citet{fifty2021efficiently} proposed a method to select sub-tasks based on task gradients.

\vspace{-0.05in}
\paragraph{Auxiliary task for adversarial learning.}
Partial sharing of model parameters is the mainstream in multi-task learning, which attempts to divide the features of different tasks into private and shared spaces. However, the shared feature space could contain some unnecessary task-specific features, while some sharable features could also be mixed in private space, suffering from feature redundancy. To alleviate this problem, \citet{liu2017adversarial} adds an adversarial task via a discriminator to estimate what task the encoding sequence comes from. Such learning strategy prevents the shared and private latent feature spaces from interfering with each other. 
This setup has also received success in multi-task multi-domain training for domain adaptation~\cite{yu2018modelling}. 
The adversarial task in this case is to predict the domain of the input. By reversing the gradient of the adversarial task, the adversarial task loss is maximized, which is beneficial for the main task as it forces the model to learn representations that are indistinguishable between domains. 

\paragraph{Auxiliary task to boost representation learning.}

While auxiliary tasks are utilized to assist the main task, they are usually expected to learn representations shared or helpful for the main task \cite{ruder2017overview}. Self-supervised, or unsupervised tasks, therefore, are often considered as a good choice.
Self-supervised objectives allow the model to learn beneficial representations without leveraging expensive downstream task labels.
For example, language modeling can help to learn transferable representations.
In BERT~\cite{devlin2019bert} pre-training, the next sentence prediction task is used to learn sentence-level representations, which is complementary to the masked language model task that mainly targets at word-level contextual representations.
Besides, \citet{rei2017semi} showed that learning with a language modeling objective improves performance on several sequence labelling tasks. 
An auto-encoder objective can also be used as an auxiliary task.  \citet{yu2020crossing} demonstrated adding the auto-encoder objective improves the quality of semantic representations for questions and answers in the task of answer retrieval.

Another branch of auxiliary tasks used to facilitate representation learning is knowledge distillation. This is achieved by forcing a smaller student model to learn a larger teacher model's output distribution or hidden representation using additional training objectives \cite{hinton2015distilling}. Knowledge contained in the hidden representations is then transferred from the teacher to the student. Thus, the student model gains the generalization ability of the teacher model, but still preserving its small size which is more suitable for deployment. Such distillation idea has been verified on popular NLP models such as BERT \cite{sanh2019distilbert}.

\subsection{Multi-step Training}


\paragraph{Narrow the search space of the subsequent decoder.}
In some cases, it is not easy to predict the original task directly due to the large search space of the potential outputs~\cite{lewis2020retrieval}.
For example, in open-domain QA, directly answering a given question is hard. So, multi-stage methods (e.g., retrieve-then-read) are often used to tackle open-domain QA problems: a retriever component finding documents that might contain the answer from a large corpus, followed by a reader component finding the answer in the retrieved documents. Documents provided by the retriever serves as conditions to the reader, which narrows the search space and thus reduces the difficulty of open-domain QA~\cite{wang2018r3,wang2018evidence}. 

In another example about pre-trained language models, BERT only learns from 15\% of the tokens that are masked in the input. ELECTRA~\cite{clark2020electra} proposed a two-step self-supervised training to improve training efficiency. The masked language modeling task performed by an auto-encoder serves as an auxiliary task. It reconstructs the masked tokens in the input. Then, a discriminative model in the second step predicts whether each token in the corrupted input was replaced by the auto-encoder. The design of such classification task allows supervision on all tokens in the example.
\paragraph{Select focused contents from the input.}
The auxiliary task can be used to focus attention on parts of the input text that can be leveraged for the main task. For example, humans tend to write summaries containing certain keywords and then perform necessary modifications to ensure the fluency and grammatical correctness of the summary. Thus, keyword extraction could help the model to focus on salient information that can be used in the summary~\cite{li2020keywords}. A similar approach can be found in \citet{cho2019mixture}, where the authors used a flexible continuous latent variable for content selection to deal with different focuses on the context in question generation.



\paragraph{Predict attributes of the output.}
In some NLG scenarios, it may be hard to guarantee the output sequence contains certain desired patterns or features (e.g., emotion, sentiment) if no explicit signals are given. Therefore, an attribute classifier could be used for predicting whether the output sequence contains the desired objective, either before or after the prediction is made. For example, \citet{fan2018question} predicted which question type should be used before generating diverse questions for an image. The predicted question type acts as an additional condition while the decoder is searching for the best question sequence. Besides, \citet{song2019generating} used a emotion classifier after the decoder to discriminate whether the generated sentence expresses the desired emotion. The post-decoder classifier guides the generation process to generate dialogue responses with specific emotions. 

\vspace{-0.05in}
\paragraph{Introduce external knowledge.}
Precisely manipulating world knowledge is extremely hard for a single neural network model. 
One could devise learning tasks informed by the knowledge so that the model is trained to acquire and utilize external knowledge. 
This research direction is known as ``Knowledge-enhanced NLP''~\cite{yu2020survey}.
The knowledge-related tasks can be combined as auxiliary to the main task, resulting in a multi-task learning setting~\cite{dinan2019wizard,kim2020sequential,zhang2021knowledge}.
For instance, \citet{wu2019duconv} uses the input sequence to query the candidate knowledge pieces via attention mechanism, then fuses the selected knowledge into decoder. The knowledge selection phase is trained by minimizing the KL-divergence between the prior distribution (queried by the input) and the posterior distribution (queried by the output).

%
%

\section{Future Directions}
\label{sec:future}

{In this section, we will discuss some promising directions regarding either task relatedness or training methods of multi-task training in NLP.}
\vspace{-0.15cm}

{\subsection{Regarding Task Relatedness}}

{\paragraph{Task-specific multi-task pre-training.} Under a typical \textit{``pre-train then fine-tune''} paradigm, many NLP works attempted to design pre-training tasks that are relevant to downstream objectives~\cite{fevry2020entities,wang2021kepler}. Such approaches endow the model with task-specific knowledge acquired from massive pre-training data. For example, \citet{wang2021kepler} learned a knowledge embedding objective besides masked language modeling (MLM) to assist relation classification and entity typing tasks; \citet{fevry2020entities} and \citet{edmem} added an entity linking objective into pre-training for fact checking and question answering applications.
These results have shown that designing proper downstream-oriented pre-training tasks is a promising direction. Such pre-training tasks are jointly trained with the MLM objective to learn relevant knowledge of downstream tasks, which can greatly reduce the gap between pre-training and fine-tuning. These tasks need to be self-supervised on pre-training corpus, while sharing a similar learning objective with downstream tasks so that relevant knowledge can be transferred.}

{\paragraph{Learning to multi-tasking.} One critical issue in MTL is how to \textit{train} a multi-task model. Existing works typically design MTL training strategies (e.g., weighing losses or task grouping) by human intuition and select the best framework through cross validation. Such model selection suffers from heavy computational cost when considering every possibility. Thus, a promising direction is to learn to multi-tasking. Meta learning is a popular approach while encountering ``learning to learn'' problems \cite{hospedales2021meta}. It aims to allow the model to quickly learn a new task, given the experience of multiple learning episodes on different tasks. \citet{wang2021bridging} tried to fuse the feature of fast adaptation of meta learning into an efficient MTL model. This approach preliminarily proved that meta-learning philosophy can benefit the training of MTL models. For future directions, using meta-learning to learn a general purpose multi-task learning algorithm is a promising route to ``learning to multi-tasking''. Besides, learning to group tasks through meta-learning is worthy of exploration.}

\vspace{-0.15cm}
{\subsection{Regarding Training Methods}}

{\paragraph{Adaptive parameter sharing.} Parameter sharing is believed to be an effective technique in improving the generalizability of multi-task learning models and reducing training time and memory footprint. Two popular parameter sharing methods are hard and soft sharing \cite{ruder2017overview}. Hard parameter sharing \cite{bekoulis2018adversarial} means all tasks share a certain number of model layers before branching out. Soft parameter sharing \cite{duong2015low} adds constraints to the distances between specific layers of different tasks. However, hard sharing suffers from finding the optimal configuration while soft sharing does not reduce the number of parameters. Therefore, in addition to empirically tuning which layers to share, learning adaptable sharing for efficient MTL is a promising solution. \citet{sun2020adashare} tried to allow adaptive sharing by learning which layers are used by each task through model training. This approach suits in the field of computer vision where many models have the architecture of stacking the same layer. However, in NLP neural networks, layers are functionally and structurally discrepant, such as the encoder-decoder framework. The development of proper adaptive sharing methods to improve parameter sharing in multi-task NLP models is needed.}

{\paragraph{Mutli-task leaning in training a universal model.}
Recently, training a universal model to perform a variety of tasks becomes an emerging trend in NLP. Multi-task supervised learning helps the model fuse knowledge from different domains, and encourages it to obtain universal representations that generalize to different downstream tasks. For example, \citet{liu2019multi} unified the input format of GLUE tasks to feed into a single model before fine-tuning on individual tasks. 
However, the role of multi-task learning is still unclear in training a universal model, with different approaches adopting MTL in different phases of transfer learning. Among recent works, \citet{aribandi2021ext5} preferred multi-task pre-training over multi-task fine-tuning for smaller gaps between pre-training and fine-tuning. \citet{aghajanyan2021muppet} used multi-task pre-finetuning on a self-supervised pre-trained model before further fine-tuning on downstream tasks. \citet{sanh2022t0} used prompted multi-task fine-tuning over a pre-trained T5~\cite{raffel2020t5} in order to perform zero-shot transfer on out-of-domain tasks. Therefore, future researches may dive deeper into maximizing the benefits of MTL in the transfer learning paradigm, including the choice of properly including MTL in pre-training or fine-tuning for better generalization. Besides, a theoretical analysis of transfer learning regarding the benefits of MTL is also desired.}

\section{Conclusions}
\label{sec:conclusion}
In this paper, we reviewed recent work on multi-task learning for NLP tasks.
According to the types of task relatedness, we categorized multi-task NLP approaches into two typical frameworks: joint training and multi-step training.
We presented the design of each framework in various NLP applications,  and discussed future directions of this interesting topic.

\section{Limitations}
Due to the space constraint, we are only able to show some prominent application scenarios of joint training and multi-step training, in which we may not cover all existing fields with multi-task approaches. For example, in dialogue systems, dialogue act recognition and sentiment recognition can be jointly trained to capture speakers' intentions. Besides, zero-shot and few-shot approaches in the multi-task setting are also interesting directions.

As for another limitation, this work is purely theoretical without any software-level implementation of the mentioned framework. In addition, we did not list the experimental results of the mentioned models on benchmark datasets because of the space limit.

\section*{Acknowledgements}

This work was supported by NSF IIS-2119531, IIS-2137396, IIS-2142827, CCF-1901059, and ONR N00014-22-1-2507. Wenhao Yu is also supported in part by Bloomberg Data Science Ph.D Fellowship. We would also like to thank Libo Qin from Harbin Institute of Technology for his valuable suggestions to this paper.

\bibliography{anthology}
\bibliographystyle{acl_natbib}

\end{document}